# 3D Distance-color-coded Assessment of PCI Stent Apposition via Deep-learning-based Three-dimensional Multi-object Segmentation

Xiaoyang Qin, Hao Huang, Shuaichen Lin, Xinhao Zeng, Kaizhi Cao, Renxiong Wu, Yuming Huang, Junqing Yang, Yong Liu, Gang Li, Guangming Ni, *Member, IEEE*

*Abstract*—Coronary artery disease poses a significant global health challenge, often necessitating percutaneous coronary intervention (PCI) with stent implantation. Assessing stent apposition holds pivotal importance in averting and identifying PCI complications that lead to in-stent restenosis. Here we proposed a novel three-dimensional (3D) distance-color-coded assessment (DccA) for PCI stent apposition via deep-learning-based 3D multi-object segmentation in intravascular optical coherence tomography (IV-OCT). Our proposed 3D DccA accurately segments 3D vessel lumens and stents in IV-OCT images, using a spatial matching network and dual-layer training with style transfer. It quantifies and maps stent-lumen distances into a 3D color space, facilitating 3D visual assessment of PCI stent apposition. Achieving over 95% segmentation precision, our proposed DccA enhances clinical evaluation of PCI stent deployment and supports personalized treatment planning.

Manuscript received XX XXXXXX 2024; revised XX XXXXXX 2024; accepted XX XXXXXX 2024. Date of publication XX XXXXXX 2024; date of current version XX XXXXXX 2024. This work was supported in part by the National Science Foundation of China under Grants 61905036, China Postdoctoral Science Foundation under Grants 2021T140090 and 2019M663465, Fundamental Research Funds for the Central Universities (University of Electronic Science and Technology of China) Under Grant ZYGX2021J012, and Medico-Engineering Cooperation Funds from University of Electronic Science and Technology of China under Grant ZYGX2021YGCX019. (Corresponding authors: Gang Li, Guangming NI.)

Xiaoyang Qin, Shuaichen Lin, Xinhao Zeng, Kaizhi Cao, Renxiong Wu, Yong Liu, and Guangming Ni are with the School of Optoelectronic Science and Engineering, University of Electronic Science and Technology of China, Chengdu 611731, China (e-mial: guangmingni@uestc.edu.cn).

Hao Huang, and Gang Li are with the Department of Cardiology, Sichuan Provincial People's Hospital, University of Electronic Science and Technology of China, Chengdu 611731, China (e-mial: ligang8252@qq.com).

Yuming Huang is with the Department of Catheterization Lab, Guangdong Cardiovascular Institute, Guangdong Provincial Key Laboratory of South China Structural Heart Disease, Guangdong Provincial People's Hospital (Guangdong Academy of Medical Sciences), Southern Medical University, Guangzhou 510085, China.

Junqing Yang is with the Department of Cardiology, Guangdong Cardiovascular Institute, Guangdong Provincial Key Laboratory of Coronary Heart Disease, Guangdong Provincial People's Hospital (Guangdong Academy of Medical Sciences), Southern Medical University, Guangzhou 510085, China.

Xiaoyang Qin, Hao Huang, and Shuaichen Lin are the co-first authors.

## I. INTRODUCTION

CORONARY artery disease (CAD) [1], [2] is a leading cause of global mortality and morbidity [3] characterized by vascular intimal contour narrowing or occlusion due to coronary atherosclerosis [4], ultimately leading to myocardial ischemia, hypoxia, or necrosis [5]. Percutaneous coronary intervention (PCI) has emerged as the primary treatment modality for most CAD cases [6], typically combined with stent implantation to alleviate vascular narrowing and reduce the risk of post-treatment vascular occlusion [7], [8]. However, inappropriate stent implantation may induce vascular tissue injury, triggering an exaggerated healing response that leads to neointimal coverage over the stent and subsequent in-stent restenosis [9], [12]. Meanwhile, stent malapposition, possibly attributed to stents undersized relative to the luminal cross-sectional area, can also culminate in in-stent restenosis and may indicate a transition from non-calcified to calcified plaques [48]. Therefore, accurate assessment of stent apposition is pivotal for preventing and promptly diagnosing in-stent restenosis, stent malapposition, and stent fracture.

Intravascular optical coherence tomography (IV-OCT) has gained widespread popularity as a high-resolution light-based imaging technique for assessing stent implantation within coronary arteries in vivo [14]. The principle of IV-OCT revolves around the interference measurement of time-delayed backscattered light from the vascular wall [10]. Metal stents strongly reflect the light emitted by the optical fiber catheter, generating distinctive bright spots and shadows in IV-OCT images [11], thereby enabling recognition by physicians. A single IV-OCT scan produces hundreds of two-dimensional (2D) cross-section images to reconstruct 3D intravascular morphology. However, IV-OCT images consisting of multiple 2D cross-section images can not directly present vessel intimal contour (lumen) and embedded stent 3D morphology, and manually analyzing those 2D cross-section images not only can't take full advantage of 3D information but also is highly time-consuming and labor-intensive [12], [13]. Therefore, automated 3D analysis is essential for achieving real-time personalized medical care, such as 3D simultaneous visualizations of stents and vessel lumens.

Before our research, numerous approaches had been reported for automatic detections of stents. Early studies primarily concentrated on manually designed strut features using 2D cross-sectional images [14], [22], such as maximum reflectance intensity in each A-line [14], etc. Subsequently, machine learning-based algorithms were proposed, encompassing approaches such as algorithms utilizing continuous wavelet transform with probability neural network [15] and fuzzy C-means [16], the K-nearest neighbor method with feature vectors [17], and bagging decision trees with feature engineering [18], [22]. Furthermore, Lu et al. further developed a support vector machine classification algorithm for assessing tissue coverage [23], while Wang et al. proposed a Bayesian network utilizing the 3D structure of stent [19].

Recently, deep learning techniques, boosting object detection and pattern recognition have also found applications in the domain of stent detection [24]. For instance, Zhou et al. introduced an automatic bioabsorbable stents (BVS) detection method based on U-net [25], and Wu et al. employed deep convolutional neural networks to detect and segment coronary artery stent struts in OCT images [12]. Additionally, Huang et al. proposed a weakly supervised fusion U-Net method based on convolutional attention mechanisms and dilated convolution to separate the contour of OCT BVS [26], while Yu et al. developed hybrid algorithms by applying standard U-Net, MobileNetV2, and DenseNet1 [20] for segmenting metal stents and BVS [27]. Furthermore, Yang et al. devised deep learning methods for analyzing the strut coverage of thin (<0.3 mm) and thick (>0.3 mm) tissues and accurately analyzed the strut area of vessels with multiple stents [11].

Although displaying promising results, aforementioned approaches have certain limitations: (1) facing difficulties in identifying stents that exhibit lower brightness or have brightness similar to the surrounding tissue during extensive data validation processes; (2) relying solely on features extracted from individual 2D IV-OCT images or a limited number of consecutive slices for training 2D CNNs neglects the overall distribution pattern of the stent in the depth direction (in 3D space); (3) above all, most studies have primarily focused on accurately segmenting stents, with less emphasis on reconstructing the entire stent and lumen in 3D space. Consequently, the intuitive and quantitative 3D assessment of stent apposition offering crucial information for real-time decision-making by physicians, is still missing.

Here we proposed a 3D distance-color-coded assessment (DccA) of PCI stent apposition via deep-learning-based 3D multi-object segmentation in IV-OCT images. Our proposed 3D DccA first achieved high-accuracy 3D segmentations of vessel lumens and PCI stents, further measured their physical distances and mapped these distances into the HVS color space to finally achieve 3D DccA of PCI stent apposition. In detail, a spatial matching network integrating 2D and 3D convolutions is proposed (inspired by 2.5D CNN [28]) to extract intra-plane and inter-plane features more effectively of 3D stents and lumens. To improve the detection of subtle stents with low brightness or tissue-like characteristics, our proposed 3D DccA introduced a fully automatic style transfer strategy based on the VGG19 network, thus generating a dataset including subtle stents. Following this, the newly generated subtle dataset is merged with the original dataset and subsequently fed into the spatial matching network for a secondary training iteration, aiming to enhance the performance of the 3D DccA. After obtaining precise 3D segmentations of vessel lumens and stents, we calculate the distance between each stent point and the lumen and then map it onto the HSV color space. This process generates 3D geometrically correlated HSV images, effectively indicating the stent apposition.

This study aims to fulfill the unmet clinical needs of PCI therapy, and our contributions are summarized as follows: (1) we propose a novel dual-layer deep learning training approach incorporating style transfer, coupled with a spatial matching network tailored for IV-OCT data, achieving more accurate 3D segmentations of stents and lumens; (2) expanding on the accurate 3D segmentations, we propose a novel method to indicate the positioning of the stent within the vessel, facilitating the prompt provision of comprehensive, intuitive, and precise information on stent-to-intimal apposition to the physicians. All 3D datasets, including the original 3D IV-OCT images (C-scans) used, 3D segmentation results, and the corresponding ground truth can be downloaded and viewed from the link: https://tianchi.aliyun.com/notebook-ai/myDataSet#datalabId=180437.

## II. METHODS

### A. Automated metallic stent and lumen analysis method

The overall workflow of our proposed 3D DccA is delineated In Fig. 1(a), 3D IV-OCT images and labels in Cartesian coordinates, serve as inputs to a 2.5D spatial matching network, primarily tasked with initial stent and lumen region detection. Subsequently, leveraging the outcomes of the initial training alongside stent label data, the areas within the stent that remain unidentified and those already identified are automatically segregated into style and content images, respectively. These are then channeled into the style transfer module for the extraction of features from the unidentified stents, which are then transferred to the identified stents, as illustrated in Fig. 1(a). The resultant dataset, enriched with challenging stent cases alongside the original images, is subsequently employed for the network's second training iteration. Finally, the segmented stent and lumen regions undergo distance-color coding before being subjected to 3D reconstruction in the HVS color space. Each of these steps will be elaborated upon further in subsequent sections.

*1) Spatial matching network for stent segmentation*: 2D CNNs, such as U-Net [31] and ResNet [32] ignore important depth information inherent in IV-OCT images (in 3D space). IV-OCT imaging operates in a spiral scanning mode, characterized by fixed and stable rotation and pullback speeds, with the stent itself possessing a regular structure. Guided by these factors, there exists a discernible distribution pattern in the depth direction for stent points within each image slice.

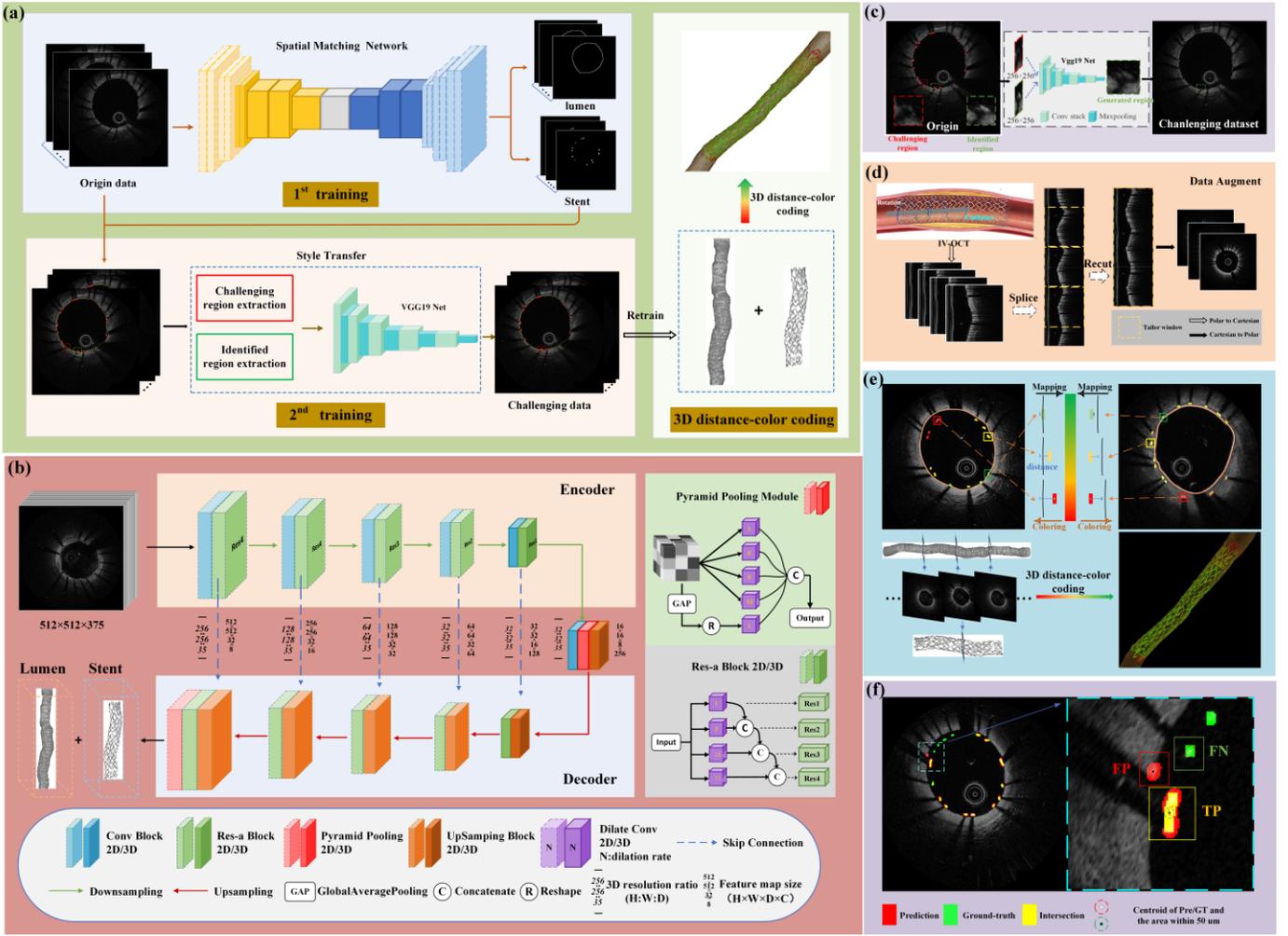

Fig. 1 The proposed method of 3D distance-color-coded PCI stent apposition reconstruction. (a) Overview of the proposed 3D DccA. (b) Proposed spatial-match encoder-decoder network for stent segmentation in IV-OCT images. (c) Style transfer strategy. (d) Data augmentation method for IV-OCT. (e) Stent apposition assessment by distance mapping to HSV color space. (f) Proposed performance evaluation diagram.

Physicians and annotation experts frequently analyze IV-OCT images by referring to adjacent frames before and after. While 3D CNNs can extract features along the depth direction of IV-OCT images, most, such as 3D Unet [33], are designed for segmenting isotropic resolution data and may not be well-suited for the specific characteristics of IV-OCT images due to the discrepancy between the in-plane resolutions (15 - 25 μm axially and 20 - 40 μm laterally) and the inter-plane resolution (typically 100 - 200 μm). Here, we proposed a spatial matching network that amalgamates both 2D and 3D convolutions. This innovative approach allows the network to address the anisotropic nature of IV-OCT image resolution while effectively leveraging both in-plane and through-plane information inherent in IV-OCT images.

We utilized the U-Net as the backbone architecture, as depicted in Fig. 1(b). To achieve consistent training as the depth of the network increases, the building blocks of the U-Net architecture were replaced with modified residual blocks of convolutional layers [34] effectively mitigating the issue of vanishing and exploding gradients commonly encountered in deep architectures [35]. The 2.5D spatial matching network comprises six levels of convolutions with skip connections between the corresponding down-sampling and up-sampling stages. Given that the original input IV-OCT images correspond to a physical volume of 7mm×7mm×75mm, while the corresponding pixel volume is 512×512×375, the first four levels exclusively employ 2D residual-a blocks to extract feature maps in 2D and encode in-plane spatial information. Subsequently, the remaining two levels utilize 3D residual-a blocks to capture both in-plane and through-plane spatial information of IV-OCT, resulting in feature maps with near-isotropic 3D resolution. Consequently, the spatial resolution of feature maps at each level stands at 256:256:35, 128:128:35, 64:64:35, 32:32:35, 32:32:35, 32:32:35, for in-plane and through-plane dimensions, respectively.

For better understanding across scales, multiple parallel atrous convolutions with different dilation rates [36] are employed within each residual-a block as shown in Fig. 1(b). The utilization of multiple parallel atrous convolutions expands the receptive field of each layer, enabling the extraction of object features across a range of receptive field scales. Four dilation rates: 1, 3, 15, and 31 were selected based on the size of the IV-OCT images and experimental evaluations. Additionally, varying numbers of parallel convolutions were employed at different layers, as depicted in Fig. 1(b).

At the bottleneck layer and the final layer, a four-level 3D Pyramid Pooling Module [37] is employed, as shown in Fig. 1(b). This module leverages a hierarchical global before integrating information from diverse scales and sub-regions, effectively fusing information across the image.

*2) Style-transfer network for challenging feature extraction*: In typical scenarios, metallic stents exhibit strong light reflection from optical fiber guides, creating a bright spot on OCT images followed by a shadow segment. Various factors, such as tissue coverage on the stent surface or residual blood surrounding the stent, can diminish the intensity of reflected light, posing challenges in identifying these areas within IV-OCT images due to their weakened features and lower occurrence rate among all stents [24].

To remedy this, a dual-layer training approach was implemented to enhance the recognition of stents with complex features. With the advancements in deep learning, neural style transfer (NST) [39] algorithms have found widespread applications in feature extraction and recombination within image processing [38]. Here the VGG-19 network was harnessed to transfer the style characteristics of challenging regions to recognized areas. By integrating the challenging dataset obtained through this process into a subsequent round of training alongside the original dataset, our 3D DccA was equipped to effectively identify stents with intricate features that were previously difficult to discern.

In the training dataset depicted in Fig. 1(c), 1196 recognized stents (similar to the one highlighted by the green box) are classified as content images, while 257 unrecognized stents (similar to the one highlighted by the red box) are labeled as style images, guided by the ground truth and masks derived from initial predictions. These stent regions, typically measuring 15 pixels × 15 pixels, undergo resizing to 256 × 256 for subsequent feature extraction and recombination using the VGG19, forming a challenging stent area for recognition, as depicted in Fig. 1(c), before being restored to their original dimensions. To ensure diversity and comprehensiveness in the features of the newly generated challenging regions, each content image undergoes random one-to-one style transfer with a style image. These resultant challenging regions overlay the original areas in the initial IV-OCT images, as demonstrated in Fig. 1(c), thereby constituting a novel challenging training dataset.

### B. Augmentation of IV-OCT images

Data augmentation [29] is a commonly adopted strategy applying rotation, translation, contrast enhancement, and noise addition to training images to improve the robustness of deep learning models. However, such typical strategies do not simply apply to IV-OCT images which are obtained in a helical manner, as illustrated in Fig. 1(d), yielding polar-coordinate images. Rotation and translation are only applicable for preprocessing data in 2D networks, as both can disrupt the inherent three-dimensional information in IV-OCT images. IV-OCT images are temporal coherence-gated depth-resolved images with mainly singly scattering signals and also with unique speckle noises due to spatial coherence [14], and uncontrolled contrast variation or noise addition may violate the underlying imaging principle and result in unrealistic images which are never possible in practice.

Here we implement a novel data-augmentation scheme, depicted in Fig. 1(d), which adheres to the specific imaging principles governing IV-OCT. This scheme not only facilitates the generation of realistic IV-OCT images but also notably enhances the model's performance. By concatenating all image sequences into a unified large 2D array, it becomes feasible to construct any valid image by employing a sliding window of identical dimensions to the original image in the rotational dimension. This sliding window technique not only enriches the diversity of 2D images for 2D networks without altering the underlying volumetric data but also enables the creation of 'new' IV-OCT pullbacks for 3D networks, where the sliding window's shifting stride matches its size. This methodology mirrors the approach utilized by Zhou et al. for coronary calcification segmentation in IV-OCT images [30].

### C. 3D distance-color-coded PCI stent apposition reconstruction

In clinical practice, physicians analyze IV-OCT images of patients undergoing PCI, examining each frame to evaluate stent-to-intimal contour apposition and formulate diagnostic and treatment strategies. Although this approach is time-consuming and labor-intensive, its biggest problem is the lack of quantifiable 3D data, such as 3D visualization. Despite attempts at 3D reconstruction post-stent segmentation in prior studies [12], quantitative insights into stent-to-vessel apposition in the reconstructed 3D images remain elusive. Additionally, these methods often fall short of effectively highlighting regions of interest.

Here we extract segmented vessel lumens and stents, and map their distance to a linear relationship within the HSV color space, as Fig. 1(e) shows. A distance threshold of 0.3 mm [11] is set as the upper limit for the mapped HSV spectrum. In cases where the neointima covers the stent by more than 0.3 mm or there is a malapposition exceeding 0.3 mm, a Hue (H) value of 0, indicating full red, is assigned. Conversely, a stent with its centroid distance to the luminal wall being 0 is assigned a Hue (H) value of 180, corresponding to full green. Stents with centroid-to-luminal-wall distances ranging from 0 to 0.3 mm are assigned Hue (H) values proportional to the distance. Fig. 1(e) depict the mapping of stent apposition at various levels in the presence of stent malapposition and tissue coverage, respectively. Ultimately, based on the aforementioned efforts to ensure the accuracy of the stent and lumen segmentation as Fig. 1(e) shows, we perform 3D reconstruction of intimal contour HSV-colored stents containing quantitative information as Fig. 1(e) shows. This enables the 3D reconstructed images to visually and quantitatively reflect the degree of fit between stents and vessel lumens.

## III. EXPERIMENTS

### A. Data Acquisition and Ground-Truth Label Generation

This study, which complied with the tenets of the Declaration of Helsinki and was approved by the Ethics Committee of the University of Electronic Science and Technology of China (ID: 10614202207190011), included data from coronary-artery-disease patients in the single-center Department of Cardiology, Sichuan Provincial People's Hospital, the affiliated hospital of the University of Electronic Science and Technology of China, with written informed consent form m all the participants.

Within the 57 IV-OCT pullbacks, cross-sections corresponding to unacceptable image quality due to incomplete flushing or non-uniform rotational distortion artifacts were excluded from the analysis, resulting in a total of 21,375 cross-sections finally analyzed. For performance testing, a total of 29 datasets were utilized, encompassing instances of well-apposition stents, malapposition, and neointimal contour coverage. Additionally, cross-sections devoid of implanted stents were incorporated into the dataset to eliminate false identifications. The manual labeling of all struts was carried out by three experienced IV-OCT analysts using ITK-SNAP software (version 3.8), followed by quality control by a senior analyst before the utilization of these labeled data.

All deep learning models were implemented in the TensorFlow v2.5.0 with NVIDIA RTX A5000 (48G) GPUs, and Python 3.8. In all comparative experiments, we assigned weights of 0.5 to the Tversky loss and BCE loss, respectively, then combined them as our loss function and selected Adam as the optimizer. For the proposed spatial-match network, we set the input data size to $512 \times 512 \times 32$, with an initial learning rate of 0.001. The same training strategy was applied to luminal segmentation..

### B. Performance evaluation

Even minor segmentation discrepancies at the pixel level can significantly impact conventional evaluation metrics, failing to adequately assess algorithmic performance. This is attributed to several factors: (1) stent struts typically exhibit dimensions of approximately 15×15 pixels with irregular shapes, posing challenges for annotators to precisely delineate ground-truth annotations based solely on stent contours. Consequently, evaluating segmentation results against pixel-level ground truth becomes impractical. (2) From a clinical perspective, achieving pixel-perfect segmentation is not essential; rather, emphasis is placed on identifying stent presence and its alignment relative to the luminal contour. Therefore, to better align with medical requirements, metal stent segmentation evaluation metrics based on precision and recall, as employed by Nam et al [21], are adopted for performance assessment.

In Fig. 1(f), three theoretical prediction scenarios are depicted, with corresponding regions of interest magnified in Fig. 1(f). Predicted results are represented by the red area, the ground truth by the green area, and the intersection between the predicted results and the ground truth by the yellow area. True positive (TP) instances occur when the centroid of a prediction falls within a 50-micrometer radius [21] of a ground truth centroid, while false positive (FP) instances arise when no ground truth centroid lies within a 50-µm radius of the prediction's centroid. Similarly, false negative (FN) instances are identified when no prediction centroid falls within a 50-µm radius of a ground truth centroid. Subsequently, these evaluation metrics are utilized to assess the performance of all models, as shown in Equations. (1-3).

$$Precision = \frac{TP}{TP+FP} \quad (1)$$

$$Dice = \frac{2*TP}{2*TP+FP+FN} \quad (2)$$

$$Iou = \frac{TP}{TP+FN+FP} \quad (3)$$

Here we compared our proposed spatial-match network with other state-of-the-art segmentation models, including U-Net, AttentionUnet [40], ResUNet [35], CENet [41], TransformerUnet [42], RetifluidNet [43], 3D U-Net and 3D ResUNet [44] employing the aforementioned three evaluation metrics. Additionally, to verify the efficacy of integrating style transfer with dual-layer training within the proposed architecture, corresponding ablation experiments were carried out across all deep learning networks.

## IV. RESULT

### A. Evaluation of stent segmentation performance and ablation experiments

Stent-segmentation results of the spatial matching network were compared with those of other established deep learning networks detailed in Table I, highlighting the superior accuracy of the 2.5D spatial matching network without style transfer. This is evidenced by its Precision (0.966±0.068), Dice (0.940±0.094), and IoU (0.900±0.137) metrics, outperforming other methods.

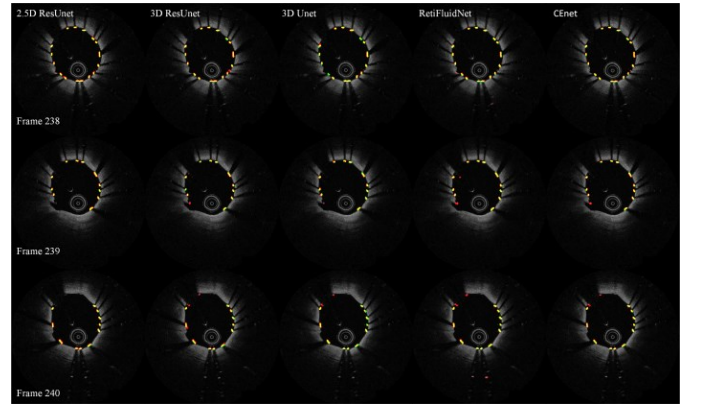

Fig. 2 Visual comparisons between state-of-the-art methods versus our proposed method. Green areas represent ground truth, red areas represent segmentation results, and yellow areas represent the overlapping regions.

Visual comparisons between the proposed 2.5D spatial matching network and other deep learning networks on three consecutive IV-OCT images are presented in Fig. 2. The 2D network tends to yield false positives in stent-like regions and overlooks less prominent features, leading to false negatives. RetiFluidNet misidentifies bright spots away from the luminal wall as stents. Although the 3D network minimizes misclassifications of stent-like morphology in implausible spatial positions, it still exhibits occasional false positives or negatives near the luminal wall. Our proposed 2.5D spatial matching network accurately captures rich 2D semantic information while effectively extracting the information in the depth direction.

TABLE I

EVALUATION RESULTS FOR STENTS SEGMENTATION AGAINST GROUND TRUTH. RESULTS SHOW A COMPARISON OF PERFORMANCE BETWEEN STATE-OF-THE-ART METHODS VERSUS OUR PROPOSED METHOD, WITH AND WITHOUT STYLE TRANSFER DUAL LAYER TRAINING STRUCTURE.

| Network | If style transfer | Segmentation Metrics | | |
|---|---|---|---|---|
| | | Precision | Dice | Iou |
| UNet | | 0.890±0.153 | 0.890±0.153 | 0.854±0.174 |
| | √ | 0.931±0.110 | 0.927±0.107 | 0.879±0.151 |
| ResUnet | | 0.898±0.149 | 0.913±0.130 | 0.860±0.168 |
| | √ | 0.934±0.109 | 0.928±0.124 | 0.881±0.154 |
| CENet | | 0.894±0.160 | 0.907±0.150 | 0.856±0.182 |
| | √ | 0.921±0.109 | 0.927±0.111 | 0.878±0.147 |
| RetiFluidNet | | 0.884±0.110 | 0.924±0.084 | 0.868±0.124 |
| | √ | 0.942±0.092 | 0.935±0.082 | 0.889±0.126 |
| TransUNet | | 0.868±0.106 | 0.918±0.084 | 0.859±0.118 |
| | √ | 0.934±0.133 | 0.927±0.136 | 0.886±0.173 |
| 3D-UNet | | 0.911±0.083 | 0.928±0.061 | 0.872±0.100 |
| | √ | 0.917±0.087 | 0.932±0.064 | 0.879±0.104 |
| 3D-ResUnet | | 0.919±0.103 | 0.927±0.091 | 0.876±0.131 |
| | √ | 0.931±0.125 | 0.928±0.117 | 0.882±0.156 |
| Our method | | **0.966±0.068** | **0.940±0.094** | **0.900±0.137** |
| | √ | **0.973±0.062** | **0.948±0.080** | **0.911±0.121** |

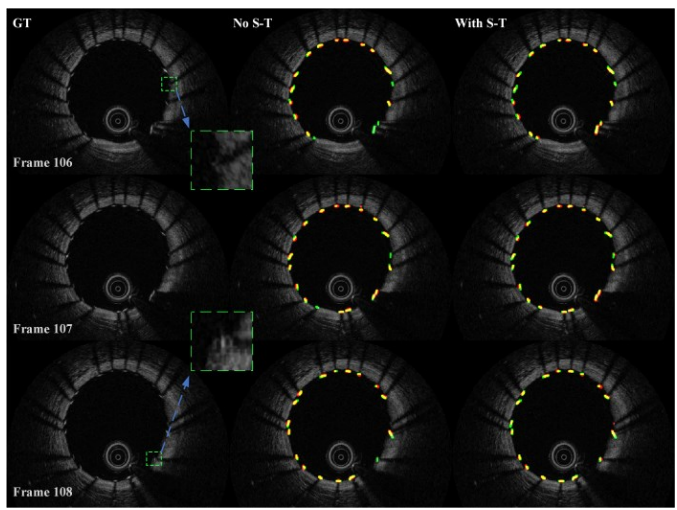

Fig. 3 Visual comparison of predictions with and without the style transfer dual-layer training structure. Green areas represent ground truth, red areas represent segmentation results, and yellow areas represent the overlapping regions.

Additionally, the efficacy of a dual-layer training structure with style transfer has been tested on various deep-learning networks, as illustrated in Table I, to ascertain its impact on stent segmentation. Incorporating this architecture notably enhances segmentation performance, particularly when confronted with the demanding dataset generated through style transfer. Among the networks evaluated, the 2D network, reliant solely on individual image information, experiences the most substantial performance boost. Notably, our proposed 2.5D spatial matching network demonstrates enhanced segmentation metrics, achieving Precision (0.973±0.062), Dice (0.948±0.080), and IoU (0.911±0.121).

Visual comparison of segmentation results with and without style transfer dual-layer training structure is depicted in Fig. 3 using three consecutive IV-OCT images. The stent regions highlighted within the green boxes in Fig. 3 exhibit characteristics such as low brightness or sparse tissue coverage, posing challenges for segmentation using conventional models. However, employing a model trained with the style transfer dual-layer training structure enables successful segmentation of these subtle stents.

B. Evaluation of lumen segmentation performance

Table II showcases the segmentation performance of 3D networks, 2D networks, and our proposed method for luminal segmentation. Interestingly, no significant disparity in segmentation performance among these networks is evident. However, it is worth noting that the segmentation performance of 2D networks marginally surpasses that of 3D networks, likely attributed to the absence of specific depth-related regularities within the lumen, unlike in stent segmentation.

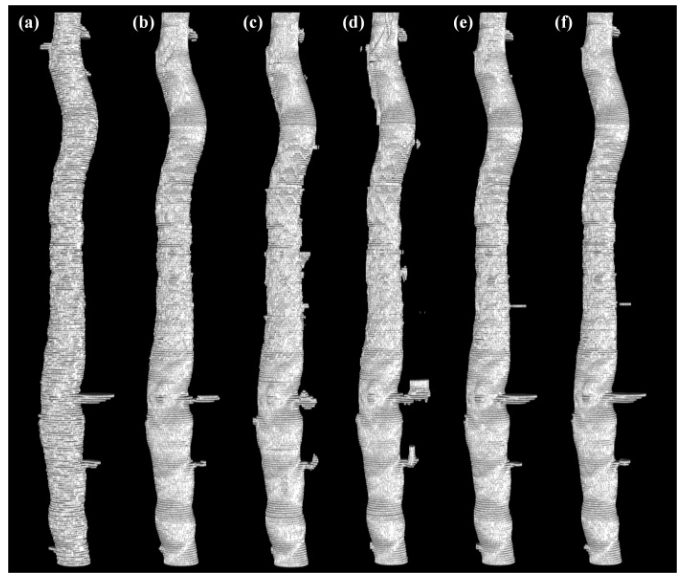

Fig. 4 3D visual comparison of lumen segmentation. (a) Ground truth, (b) our proposed method, (c) 3D ResUnet, (d) 3D Unet, (e) RetiFluidnet, and (f) Unet are displayed.

Upon examining the 3D reconstructed images depicting the segmentation results of each network in Fig. 4, it becomes apparent that the contour morphology of vascular bifurcation branches is somewhat compromised by the 3D network, in comparison to the 2D networks. In contrast, our proposed 2.5D spatial matching network, amalgamating the characteristics of both 2D and 3D networks, effectively preserves a certain degree of branch morphology while circumventing the occurrence of small-volume impurities often present in 2D networks.

TABLE II
EVALUATION RESULTS FOR LUMEN SEGMENTATION AGAINST GROUND TRUTH. RESULTS SHOW A COMPARISON OF PERFORMANCE BETWEEN STATE-OF-THE-ART METHODS VERSUS OUR PROPOSED METHOD.

| Network | Precision | Dice | IOU | Jaccard | Recall |
|---|---|---|---|---|---|
| UNet | 97.36±1.08 | 97.98±0.98 | 95.73±1.77 | 96.05±1.83 | 98.64±2.00 |
| ResUnet | 96.63±1.62 | 97.54±1.97 | 94.90±3.41 | 95.26±3.27 | 98.57±3.40 |
| CENet | 96.75±1.41 | 97.63±0.99 | 95.07±1.83 | 95.38±1.85 | 98.57±2.09 |
| RetiFluidNet | 96.95±1.19 | 97.77±1.01 | 95.30±1.82 | 95.65±1.87 | 98.64±2.04 |
| TransUNet | 95.80±1.66 | 97.34±0.92 | 94.50±1.69 | 94.84±1.71 | 98.98±1.67 |
| 3D-UNet | 96.33±1.55 | 97.27±1.57 | 94.42±2.71 | 94.72±2.74 | 98.31±3.03 |
| 3D-ResUnet | 95.32±1.84 | 96.61±2.16 | 93.15±3.80 | 93.52±3.81 | 98.10±4.39 |
| Our method | 97.80±1.14 | 97.76±1.42 | 94.93±2.43 | 95.66±2.52 | 97.78±2.69 |

## C. 3D distance-color coded PCI stent apposition Assessment

The stent and lumen segmentation outcomes generated by 2D, 3D networks, and our proposed method underwent evaluation for stent apposition, employing the 3D distance-color-coded method outlined in Section II.C, illustrated in Fig. 5.

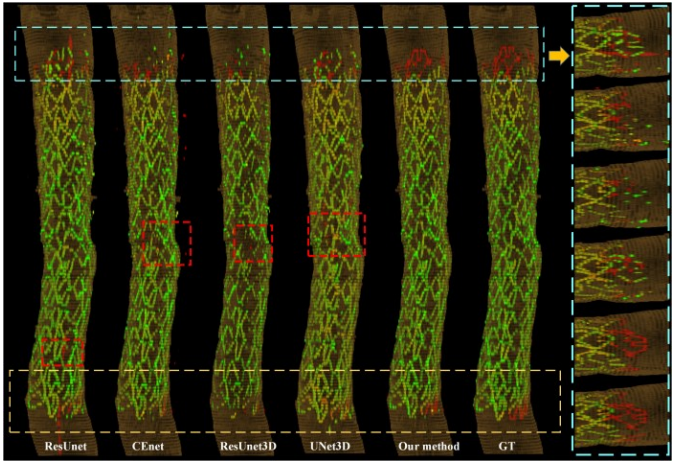

Fig. 5 3D distance-color-coded assessment of PCI stent deployment utilizing segmentation results. The yellow box, blue box, and red box refer to two regions with noteworthy differences, with the blue box enlarged to the right for better comparative observation.

Within regions highlighted by blue and yellow boxes in Fig. 5, indicating suboptimal stent apposition, our proposed spatial matching network exhibits superior extraction of the stent structures compared to both 2D and 3D networks, achieving a more comprehensive and cleaner delineation. While both 2D and 3D networks demonstrate a tendency to misclassify well-apposition areas, as evidenced by the presence of a red box in Fig. 5, misclassifications by 3D networks are primarily concentrated near the intimal contour, whereas those by 2D networks are distributed both within and outside the intimal contour. Notably, the 3D network occasionally misjudges regions proximal to the stent intimal contour. In contrast, the 2D network sporadically generates impurity points far from the intimal contour, even erroneously classifying catheters as inadequately apposition stent points. Our 2.5D spatial matching network showcases minimal misjudgments in well-apposition regions compared to 2D and 3D networks.

Table III displays the stent segmentation performance on IV-OCT data, encompassing tissue coverage and malapposition cases, for the 3D network, 2D network, and our proposed method. Notably, our proposed method demonstrates superior metrics in the malapposition group (Precision: 0.955±0.069, Dice: 0.929±0.072, IoU: 0.875±0.117) and the tissue coverage group (Precision: 0.954±0.088, Dice: 0.919±0.089, IoU: 0.862±0.141) compared to both the 2D and 3D networks.

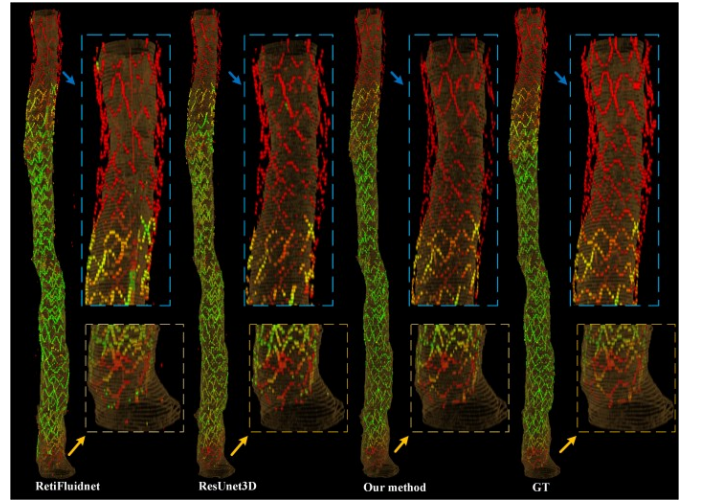

Fig. 6 3D distance-color-coded assessment of IV-OCT data covering both intimal coverage and malapposition. The blue box corresponds to the region where the stent is covered by intimal, while the yellow box corresponds to the region of stent malapposition

The segmentation results of the 2D and 3D networks, along with our proposed method, on IV-OCT data featuring both tissue coverage and malapposition, are visually presented using the 3D distance-color-coded approach, as demonstrated in Fig. 6. These images reveal that our proposed method's segmentations offer a more comprehensive and accurate representation in regions where the stent is covered by tissue, as depicted by the blue box in Fig. 6, outperforming both the 2D and 3D networks. Similarly, the 3D distance-color-coded images derived from our proposed method's segmentations exhibit high fidelity with notably reduced misclassification rates in regions of stent malapposition, as indicated by the yellow box in Fig. 6, in comparison to both the 2D and 3D networks.

## D. Clinical Feasibility Study t

We evaluated the clinical viability of our 3D DccA approach through automated analysis of IV-OCT images from multiple patients who underwent PCI and stent implantation. In Fig. 7 - 10, we present the IV-OCT pullback images alongside the 3D DccA visualization results following automated analysis and reconstruction for four patients with varying degrees of stent-intimal contour apposition.

In the majority of patient IV-OCT images shown in Fig. 7,

TABLE III
EVALUATION RESULTS FOR STENTS EXHIBITING MALPOSITION AND COVERED BY TISSUE. RESULTS DEMONSTRATE A COMPARISON OF THE SEGMENTATION
PERFORMANCE BETWEEN THE STATE-OF-THE-ART METHOD AND THE PROPOSED METHOD.

| Network | Malapposition | | | Intimal coverage | | |
|---|---|---|---|---|---|---|
| | Precision | Dice | Iou | Precision | Dice | Iou |
| UNet | 0.909±0.103 | 0.892±0.092 | 0.816±0.140 | 0.900±0.130 | 0.851±0.144 | 0.765±0.194 |
| ResUnet | 0.936±0.088 | 0.896±0.098 | 0.826±0.147 | 0.900±0.119 | 0.853±0.133 | 0.765±0.187 |
| CENet | 0.946±0.079 | 0.899±0.099 | 0.830±0.152 | 0.923±0.112 | 0.872±0.135 | 0.795±0.190 |
| RetiFluidNet | 0.948±0.088 | 0.903±0.102 | 0.837±0.150 | 0.915±0.111 | 0.886±0.103 | 0.809±0.149 |
| TransformerUNet | 0.913±0.093 | 0.903±0.085 | 0.833±0.130 | 0.916±0.106 | 0.874±0.123 | 0.796±0.180 |
| 3D-UNet | 0.949±0.077 | 0.918±0.081 | 0.859±0.130 | 0.909±0.111 | 0.889±0.109 | 0.815±0.153 |
| 3D-ResUnet | 0.924±0.097 | 0.908±0.086 | 0.842±0.136 | 0.951±0.075 | 0.909±0.094 | 0.846±0.142 |
| **Our method** | **0.955±0.069** | **0.929±0.072** | **0.929±0.072** | **0.954±0.088** | **0.919±0.089** | **0.862±0.141** |

a close resemblance to Fig. 7(b) signifies excellent apposition between the stent and vessel intimal contour. Fig. 7(a) and Fig. 7(c) depict segments of vessel branching that are easily overlooked in the original IV-OCT images due to their limited frames. By highlighting these segments in red within the stent and vessel intimal contour, precise localization of the branching segments can be swiftly and accurately achieved. Furthermore, the 3D DccA visualization results offer a more intuitive and efficient assessment of the stent-vessel apposition, enhancing the overall clinical evaluation process.

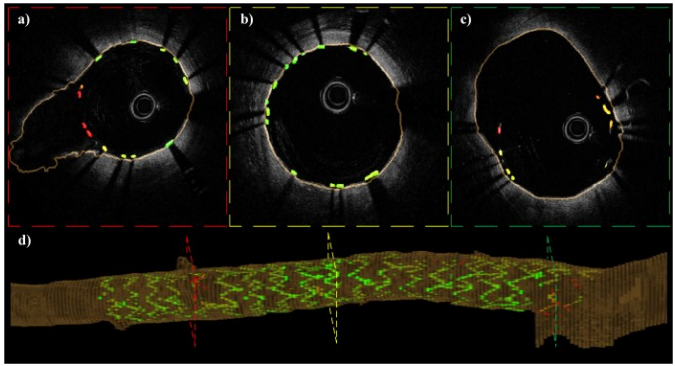

Fig. 7 3D distance-color-coded PCI stent apposition assessments for a patient with good stent apposition. (a), (b), and (c) correspond to the three original IV-OCT image slices taken from the red, yellow, and green cross-sections in d), respectively. d) represents our 3D DccA visualization result.

Fig. 8 presents an IV-OCT pullback of a patient with partial stent malapposition. Within the IV-OCT images, segments displaying well apposition (as illustrated in Fig. 8 (b)) coexist with segments demonstrating malapposition (depicted in Fig. 8 (a) and Fig. 8 (c). Determining the relative position and exact degree of malapposition based solely on the original IV-OCT images is challenging. However, our 3D DccA visualization image offers clear and intuitive insights, revealing two segments, with lengths of 9 mm and 4 mm, respectively, both exhibiting malapposition exceeding 0.3 mm.

In Fig. 9, the IV-OCT image indicates the development of neointimal contour coverage over the stent. Within the IV-OCT images, there are segments that apposition well (as depicted in Fig. 9 (a)) and segments exhibiting neointimal contour coverage (as shown in Fig. 9(b) and Fig. 9(c)). Acquiring the degree and length of neointimal contour coverage over the stent from IV-OCT images alone is challenging. Our 3D DccA visualization image offers a comprehensive view, revealing the transition of stent apposition in this vessel segment from healthy to neointima-covered. The neointima's thickness exceeds 0.3 mm, extending approximately 15 mm along the stent.

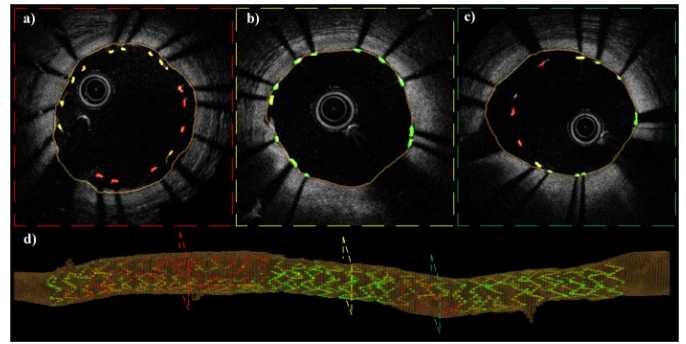

Fig. 8 3D distance-color-coded PCI stent apposition assessments for a patient with stent malapposition. (a), (b), and (c) correspond to the three original IV-OCT image slices taken from the red, yellow, and green cross-sections in (d), respectively. (d) represents our 3D DccA visualization result.

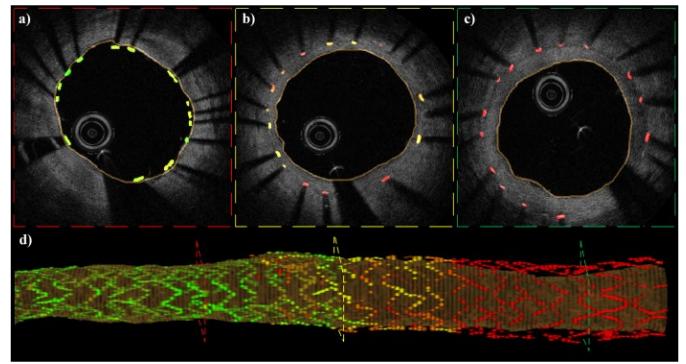

Fig. 9 3D distance-color-coded PCI stent apposition assessments for a patient with stent coverage by neointima. (a), (b), and (c) correspond to the three original IV-OCT image slices taken from the red, yellow, and green cross-sections in (d), respectively. (d) represents our 3D DccA visualization result.

Some patients exhibit severe in-stent restenosis in clinical practice, as depicted in Fig. 10. The images reveal regions of severe in-stent restenosis, indicated by the blue areas in Fig. 10 (b), where the stent features are nearly obscured by excessive neointimal contour coverage. Even experienced physicians find it challenging to identify these stents, and automated analysis methods struggle in the absence of distinctive features.

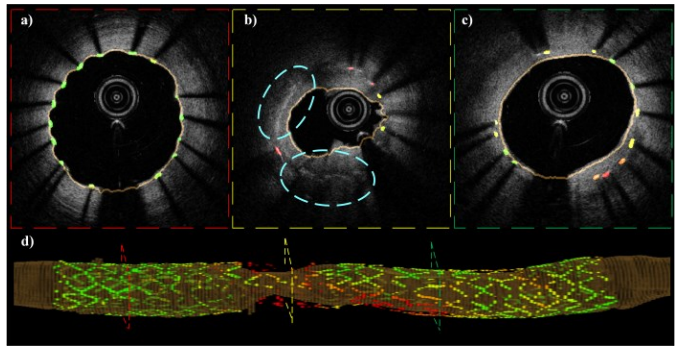

Fig. 10 3D distance-color-coded PCI stent apposition assessments for a patient with severe in-stent restenosis due to very thick neointima. (a), (b), and (c) correspond to the three original IV-OCT image slices taken from the red, yellow, and green cross-sections in (d), respectively. (d) represents our 3D visualization result. The region adjacent to the yellow slice exhibits near invisibility due to excessive neointimal contour coverage over the stent.

However, our 3D DccA visualization image provides a straightforward understanding: (1) the head and tail segments of the implanted stent apposition well, similar to Fig. 10(a); (2) a specific segment corresponding to Fig. 10(c) shows an 8-mm long visible stent covered by neointima; (3) another segment of the vessel exhibits severe stenosis, approximately 6 mm in length, with the stent substantially covered by a thick layer of neointima. Physicians should carefully consider this area and develop corresponding strategies.

The above clinical cases demonstrate the capability of our 3D DccA for PCI stent apposition assessment to provide accurate, rapid, and intuitive visualization of the apposition relationship between the vessel intimal contour and stent. The automated analysis of each IV-OCT data typically takes an average of 70 seconds. This functionality empowers physicians to swiftly gain precise insights into the specific vascular health status of individual patients, thereby expediting the formulation of personalized treatment strategies.

## V. Discussion

In-stent restenosis (ISR) is defined as luminal narrowing with >50% diameter stenosis of a stented coronary segment at follow-up angiography [45]. The prevalence of ISR varies, estimated at 5% to 10% for patients with second-generation DES [46] and requiring treatment in 10% to 20% of patients receiving DES, leading to recurrent in-stent restenosis (R-ISR) [47]. Therefore, ISR remains a significant clinical challenge for long-term patient management, emphasizing the importance of accurate evaluation and individualized treatment. Our proposed 3D DccA offers an automated and intuitive 3D display of the stent-luminal apposition, bridging functional gaps in current clinical tools. Despite potential partial stent loss in cases of excessive tissue coverage, mainly due to thick tissue approaching the OCT detection limit, manual annotation faces significant challenges. For research purposes, our 3D DccA provides a more comprehensive and intuitive presentation of stent morphology and apposition compared to manual analysis, crucial for rapidly evaluating and personalizing ISR treatment in the era of big data, thus optimizing clinical outcomes.

This research also presents several limitations. Firstly, our moderately sized dataset may benefit from an increase in the number of cases to improve the model's generalization, particularly in cases involving unwashed blood that remains analyzable manually and in segments of blood vessels with multiple stents implanted over time. Secondly, our proposed method relies on the manual annotation of thousands of IV-OCT images, a laborious and time-consuming process that demands specialized knowledge and skills. Future work could focus on developing a semi-supervised or unsupervised learning framework, aiming to learn from a limited number of labeled IV-OCT images or even unlabeled ones. Additionally, images generated by IV-OCT devices from different brands may exhibit distinct characteristics, warranting efforts to enhance the model's capability to automatically analyze images from various devices.

## VI. Conclusion

We propose a fully automated 3D distance-color-coded method for evaluating the apposition of stents and intimal contours using a spatial matching convolutional neural network, along with a style transfer dual-layer training structure based on a few distinctive features of stents in IV-OCT images. This approach significantly improves our 3D DccA' ability to recognize subtle stents. Our proposed 3D DccA accurately segments stent and intimal contours, quantifies apposition information between the stent and the intimal surface, and visualizes the information in a more intuitive 3D distance-color-coded image format. All in all, our 3D DccA can be utilized for comprehensive 3D assessment of stent apposition to the intimal surface and for developing personalized follow-up treatments.